%% file: main_arxiv.tex
\documentclass[10pt,twocolumn,letterpaper]{article}

\usepackage[pagenumbers]{wacv} 

\usepackage{graphicx}
\usepackage{amsmath}
\usepackage{amssymb}
\usepackage{booktabs}
\usepackage[accsupp]{axessibility}
\usepackage{subfiles}
\usepackage{makecell}
\usepackage{multirow}
\usepackage{tabularx}
\usepackage{enumitem}

\usepackage[pagebackref,breaklinks,colorlinks]{hyperref}

\newcommand{\Paragraph}[1]{\vspace{0.5mm} \noindent \textbf{#1} \vspace{0mm}}
\newcommand{\Section}[1]{\vspace{-1mm} \section{#1} \vspace{-0mm}}
\newcommand{\SubSection}[1]{\vspace{-0mm} \subsection{#1} \vspace{-0.5mm}}
\newcommand{\SubSubSection}[1]{\vspace{-2mm} \subsubsection{#1} \vspace{-0mm}}

\DeclareMathOperator*{\argmin}{arg\,min}

\usepackage[capitalize]{cleveref}
\crefname{section}{Sec.}{Secs.}
\Crefname{section}{Section}{Sections}
\Crefname{table}{Table}{Tables}
\crefname{table}{Tab.}{Tabs.}

\def\wacvPaperID{641} 
\def\confName{WACV}
\def\confYear{2024}

\begin{document}


\title{FarSight: A Physics-Driven Whole-Body Biometric System \\ at Large Distance and Altitude}


\author{Feng Liu$^1$, Ryan Ashbaugh$^1$, Nicholas Chimitt$^5$, Najmul Hassan$^4$, Ali Hassani$^3$, Ajay Jaiswal$^2$, \\
Minchul Kim$^1$, Zhiyuan Mao$^5$, Christopher Perry$^1$, Zhiyuan Ren$^1$, Yiyang Su$^1$, Pegah Varghaei$^1$, \\
Kai Wang$^3$, Xingguang Zhang$^5$, Stanley Chan$^5$, Arun Ross$^1$, Humphrey Shi$^3$, Zhangyang Wang$^2$, \\
\vspace{2mm}
Anil Jain$^1$ and Xiaoming Liu$^1$\\
$^1$ Michigan State University, East Lansing MI 48824, USA\\
$^2$ University of Texas at Austin, Austin TX 78712, USA\\ 
$^3$ Georgia Tech, Atlanta GA 30332, USA \\
$^4$ University of Oregon, Eugene OR 97403, USA\\
$^5$ Purdue University, West Lafayette IN 47907, USA\\\\
}

\maketitle


\subfile{sec_0_abstract.tex}

\subfile{sec_1_intro.tex}

\subfile{sec_2_related_work.tex}

\subfile{sec_3_method.tex}

\subfile{sec_4_exp.tex}

\subfile{sec_5_conclusion.tex}

{\small
\bibliographystyle{ieee_fullname}
\bibliography{egbib}
}

\end{document}

%% file: sec_0_abstract.tex
\begin{abstract}


Whole-body biometric recognition is an important area of research due to its vast applications in law enforcement, border security, and surveillance. This paper presents the end-to-end design, development and evaluation of FarSight, an innovative software system designed for whole-body (fusion of face, gait and body shape) biometric recognition. FarSight accepts videos from elevated platforms and drones as input and outputs a candidate list of identities from a gallery. The system is designed to address several challenges, including (i) low-quality imagery, (ii) large yaw and pitch angles, (iii) robust feature extraction to accommodate large intra-person variabilities and large inter-person similarities, and (iv) the large domain gap between training and test sets. FarSight combines the physics of imaging and deep learning models to enhance image restoration and biometric feature encoding. We test FarSight's effectiveness using the newly acquired IARPA Biometric Recognition and Identification at Altitude and Range (BRIAR) dataset. 
Notably, FarSight demonstrated a substantial performance increase on the BRIAR dataset, with gains of $+11.82\%$ Rank-20 identification and $+11.30\%$ TAR@1\% FAR.

\end{abstract}


%% file: sec_1_intro.tex
\section{Introduction} 


The aim of whole-body biometric recognition is to develop a person recognition system that will surpass the performance of state-of-the-art (SoTA) recognition of the face, gait, and body shape alone, specifically in the challenging, unregulated conditions present in full-motion videos (\emph{e.g.}, aerial surveillance).
It encompasses functionalities such as person detection, tracking, image enhancement, the mitigation of atmospheric turbulence, robust biometric feature encoding, and multi-modal fusion and matching. The wide-ranging applications of whole-body recognition in fields like law enforcement, homeland security and surveillance, further underscore its importance~\cite{gong2011person,zheng2016person,ross2006handbook,ross2019some}.


\begin{figure}[t]
\centering
\includegraphics[trim=0 0 0 0,clip, width=\linewidth]{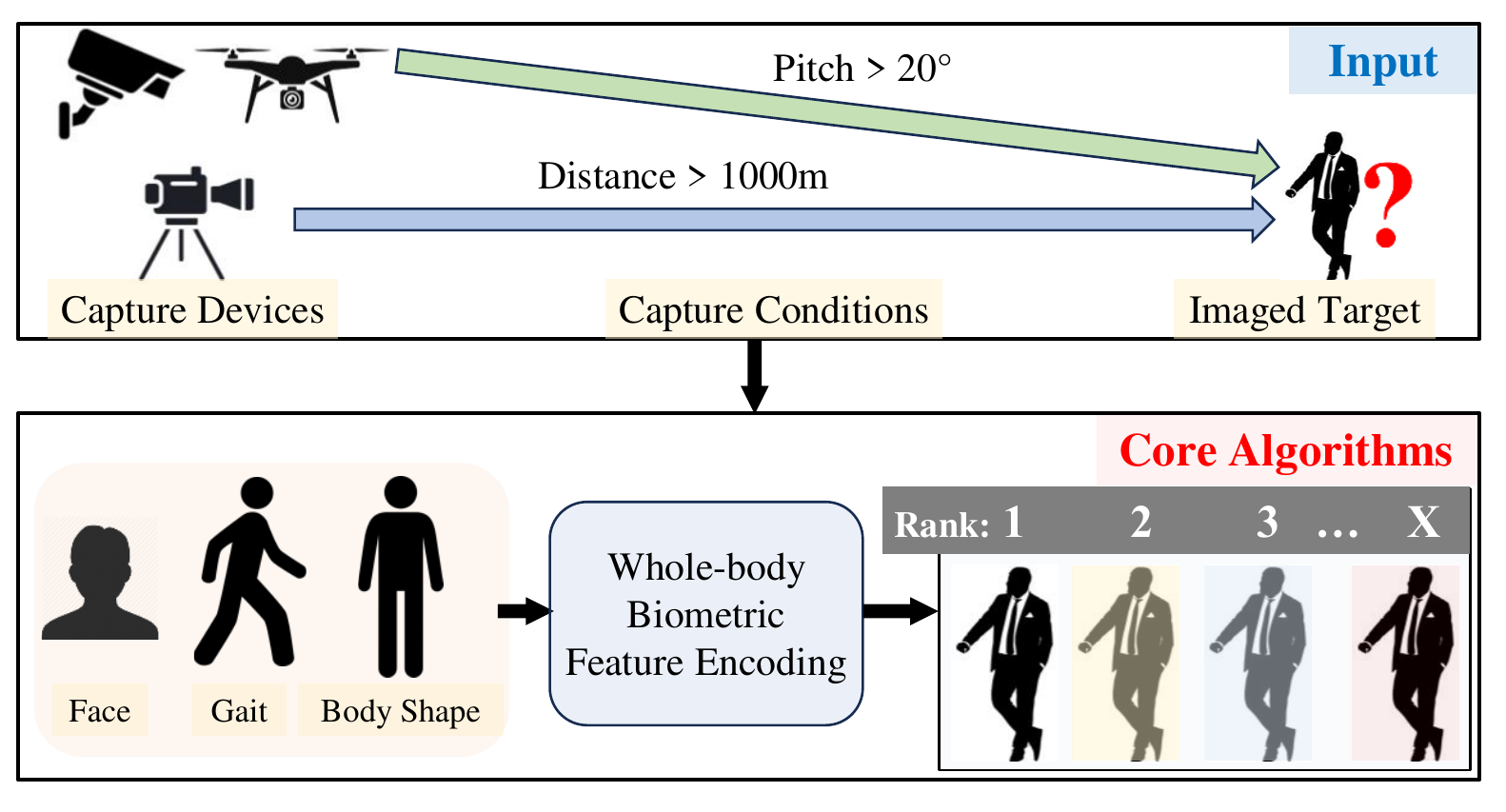}
\vspace{-8mm}
\caption{\small \textbf{FarSight} is a person recognition system that implements and fuses SoTA face, gait and body shape recognition modules in challenging conditions presented by full-motion videos. } 
\label{fig:teaser}
\vspace{-4mm}
\end{figure}

To achieve these goals, we design, prototype and evaluate a software system called \textbf{FarSight} for whole-body (face, gait and body shape) biometric recognition. 
As illustrated in Fig.~\ref{fig:teaser}, FarSight accepts as input a video captured at long-range and from elevated platforms, such as drones, and outputs a candidate list of identities present in the input video.

The design of FarSight confronts a number of novel challenges that have not been adequately addressed in existing literature: 
i) Low-quality video frames due to long-range capture (hundreds of meters) and atmospheric turbulence (with the refractive index structure parameter $C_n^2$ in ranges of $10^{-17}$ to $10^{-14}$ m$^{-2/3}$ \cite{Schmidt_2010_a}).
ii) Large yaw and pitch angles ($>20$ degrees) due to elevated platforms (altitudes of up to $400$m).
iii) Degraded feature sets due to low visual quality (the pixel range for Inter-Pupillary Distance is around $15{-}100$).
iv) Limited domain and paucity of training data due to diversity in the operating environments resulting in a large domain gap between training and test sets.


To address these challenges, the design of FarSight heavily relies on modeling the \emph{underlying physics} of image formation, image degradation and human body models throughout the recognition pipeline. Further, we integrate the learned physics knowledge into the deep learning models for feature encoding. 
The four key modules of FarSight are 1) image restoration, 2) detection and tracking, 3) biometric feature encoding, and 4) multi-modal fusion. 

\begin{itemize}
\vspace{-2mm}
\item  

Image restoration: Video streams captured from long distances suffer from atmospheric turbulence, platform vibration, and systematic aberrations. Unlike most SoTA approaches that rely on deep learning, we directly model the physics of turbulence. This model not only provides better understanding of imaging limits and turbulence parameters but also enables the creation of datasets for training restoration modules. Consequently, our approach ensures improved explainability and requires fewer labeled samples, leading to superior generalization in unseen environments.

\vspace{-2mm}
\item 
Detection and tracking:  We develop a joint body and face detection module, which is able to
associate face and body bounding boxes.
Detected bounding boxes can then be fed into an appropriate feature extractor (embedding) 
without requiring a post-processing stage to match face and body bounding boxes.

\vspace{-2mm}
\item 
Biometric (face, gait and body shape) feature encoding.
(i) Face: We leverage adaptive loss function, two-stage feature fusion, and controllable face synthesis models to effectively manage image quality variation, frame-level feature consolidation, and domain gap.
(ii) 
 Gait: We extract both local features and global correlations to improve identification in diverse scenarios.
(iii) Body shape: We learn a robust $3$D shape representation that is invariant to clothing and body pose variations, leading to improvements in body matching.


\vspace{-2mm}
\item 
Multi-modal fusion: This module performs score-level fusion and score imputation in case of missing data (when no features could be extracted for one or more biometric modalities), which does occur due to the challenging nature of long range and high angle of inclination videos.

\end{itemize}

The innovations of \textbf{FarSight} system are as follows:

$\diamond$ Explicitly modeling the physics of imaging through turbulence and image degradation and integrating physics-based models into deep learning for image restoration.

$\diamond$ Utilizing a joint body and face detection approach, easily integrated with upstream and downstream tasks.

$\diamond$ An effective feature encoding for face, gait and body shape, along with a novel multimodal feature fusion approach, enabling superior recognition performance.


$\diamond$ Utilizing the Biometric Recognition and Identification at Altitude and Range (BRIAR) dataset~\cite{cornett2023expanding}, we demonstrate the superior performance of the proposed FarSight system, and its robustness and effectiveness in whole-body biometric recognition under challenging conditions.


%% file: sec_2_related_work.tex
\begin{figure*}[t]
\begin{center}
\includegraphics[width=0.9\linewidth]{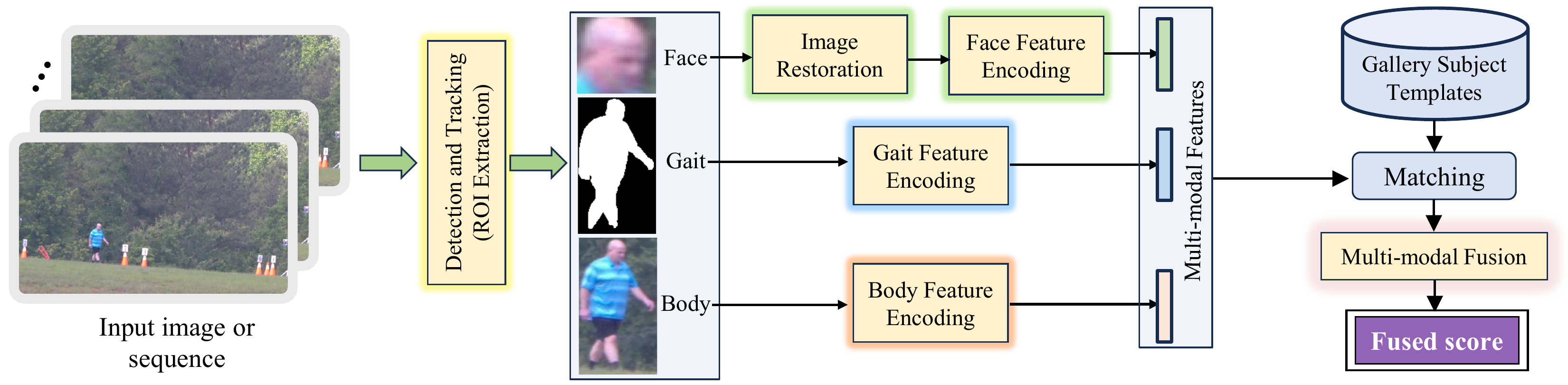}
\end{center}
\vspace{-6mm}
   \caption{\small The proposed FarSight system incorporates six components: \emph{detection and tracking, image restoration, face, gait, and body shape feature extraction, and multi-modal biometric fusion}. \vspace{-4mm}}
\label{fig:overview}
\end{figure*}
\section{Related Work}\label{sec:priors}

\Paragraph{Whole-Body Biometrics Recognition.}
Whole-body biometric recognition merges multiple physical traits, specifically face, gait, and body shape, to bolster identification accuracy, especially in challenging scenarios. Unlike traditional biometric systems  focusing on a single trait~\cite{arcface,kim2022adaface,gaitbase,CAL,connor2018biometric,zhang2020learning,liu2018disentangling,improving-face-recognition-from-hard-samples-via-distribution-distillation-loss,fan-feature-adaptation-network-for-surveillance-face-recognition-and-normalization}, this comprehensive approach can mitigate inherent weaknesses and exploit the strengths of each individual trait, leading to enhanced recognition performance.
For example, while face recognition might struggle with varying poses and lighting, gait can be affected by walking speed and attire. Body shape remains a consistent identifier, though it can vary with clothing and posture.
%
Recent literature~\cite{guo2023multi,jin2022cloth} have increasingly embraced this multi-faceted approach, but many do not provide comprehensive solutions that include image restoration, detection, tracking, and fusion of modalities.
%
This gap indicates potential for further development in holistic biometric systems, ensuring robust recognition in challenging video conditions.


\Paragraph{Physics Modeling of Imaging through Turbulence.}  
Turbulence is modeled as a stochastic phenomenon with its modern form largely based on Kolmogorov \cite{Kolmogorov_1941_a}. 
The atmosphere can be modeled as a turbulent volume that perturbs light propagating through it \cite{Tatarski_1967_a, Roggemann_1996_a}.
%
%
Since the atmosphere is a stochastic phenomenon, its effect on an image is also stochastic. Drawing realizations from this distribution requires a simulator. Simulating these effects most often comes in the form of mirroring nature: a wave is numerically propagated through a simulated atmosphere. Methods that utilize numerical wave propagation in this manner are referred to as split-step propagation \cite{Roggemann_2012_a, hardie2017simulation, Schmidt_2010_a, Hardie_2022_a}. 
%
Alternative methods combine empirical understanding and analysis~\cite{Repasi_2008_a, Repasi_2011_a, Leonard_2012_a, Potvin_2011_a} with some recent modification and improvement \cite{Miller_2019_a, Miller_2021_a}. 
Given the scarcity of open-source tools, we introduce a unique modeling approach.

\Paragraph{Image Restoration. } 
Successful biometric recognition relies upon robust feature extraction from sensed imagery~\cite{Jain_2016_a}. With poor-quality imagery, image restoration serves as a way to extract robust and salient features and potentially boost recognition accuracy. However, restoration methods may \emph{change} the person's identity based on reconstructed features as shown in attack-based work~\cite{Mai_2019_a}. Thus, reconstruction in this biometric context is slightly different. We prefer a reconstructed image that improves downstream recognition performance. Face deblurring in the presence of invariant blur has been shown to have positive results on downstream classification \cite{Shen_2018_a}. Furthermore, some efforts in restoration~\cite{Lau_2020_a, Nair_2023_a,yasarla2022cnn} have suggested that reconstruction may indeed help in the case of atmospheric turbulence degraded images. These methods, however, rely only on single frames, therefore, in the FarSight system we use multi-frame fusion to improve the quality of degraded images.

\Paragraph{Detection and Tracking.} 
Face detection has been extensively studied in the field of computer vision, with numerous endeavors aimed at detecting faces across a diverse array of scenes. Various methodologies, as presented in~\cite{li2019dsfd, zhu2020tinaface, deng2020retinaface}, have successfully employed different approaches for detecting faces in unconstrained settings.
Building upon this, pedestrian tracking is another significant module in biometrics. A multitude of strategies have been developed to improve both the efficiency and effectiveness of tracking. Among them, tracking by detection paradigms has emerged as the leading approach due to its adaptability and superior performance.
Motion-based methods~\cite{bewley2016simple, zhang2022bytetrack, zhou2020tracking} employ spatiotemporal information to enhance object association and improve tracking accuracy.
Appearance-based methods~\cite{wang2021multiple, wojke2017simple, yu2022towards}  introduce various appearance features to facilitate accurate object matching.

\Paragraph{Multi-Modal Biometric Fusion. } 
Fusion relies on leveraging encoded biometric features or scores from multiple matchers. An example of a score-level fusion method is the sum rule, where normalized scores are weighted and summed to generate the fused score to be used for performance evaluation~\cite{ross2003information, he2010performance}.

%% file: sec_3_method.tex
\Section{FarSight: System Architecture}\label{sec:}

\SubSection{Overview of FarSight}
As illustrated in Fig.~\ref{fig:overview}, FarSight operates through six modules: detection and tracking, image restoration, face, gait, and body shape feature extraction, and multi-modal fusion. These modules work within a scalable testing framework, optimizing GPU usage via adaptable batch sizes.
An API utility facilitates communication between the framework and external systems, transmitting video sequences from configuration files to the framework via Google RPC calls. Essential features extracted from these sequences are stored in HDF5 files for performance evaluation.

%

The workflow starts with input video sequences undergoing detection and tracking. Regions of interest (RoI) are identified and forwarded to gait and body modules, with face images undergoing restoration. Gait and body modules produce unique feature vectors via average pooling, while the face module, using CAFace~\cite{caface}, consolidates features across sequences. A probe comprises a single video segment per subject, while gallery enrollments – multiple video sequences and stills – are merged into a singular feature vector for each modality.

\SubSection{Challenges in FarSight}

%
%

\begin{figure}[t]
\begin{center}
\includegraphics[trim={0 0 0 7.2cm},clip, width=0.95\linewidth]{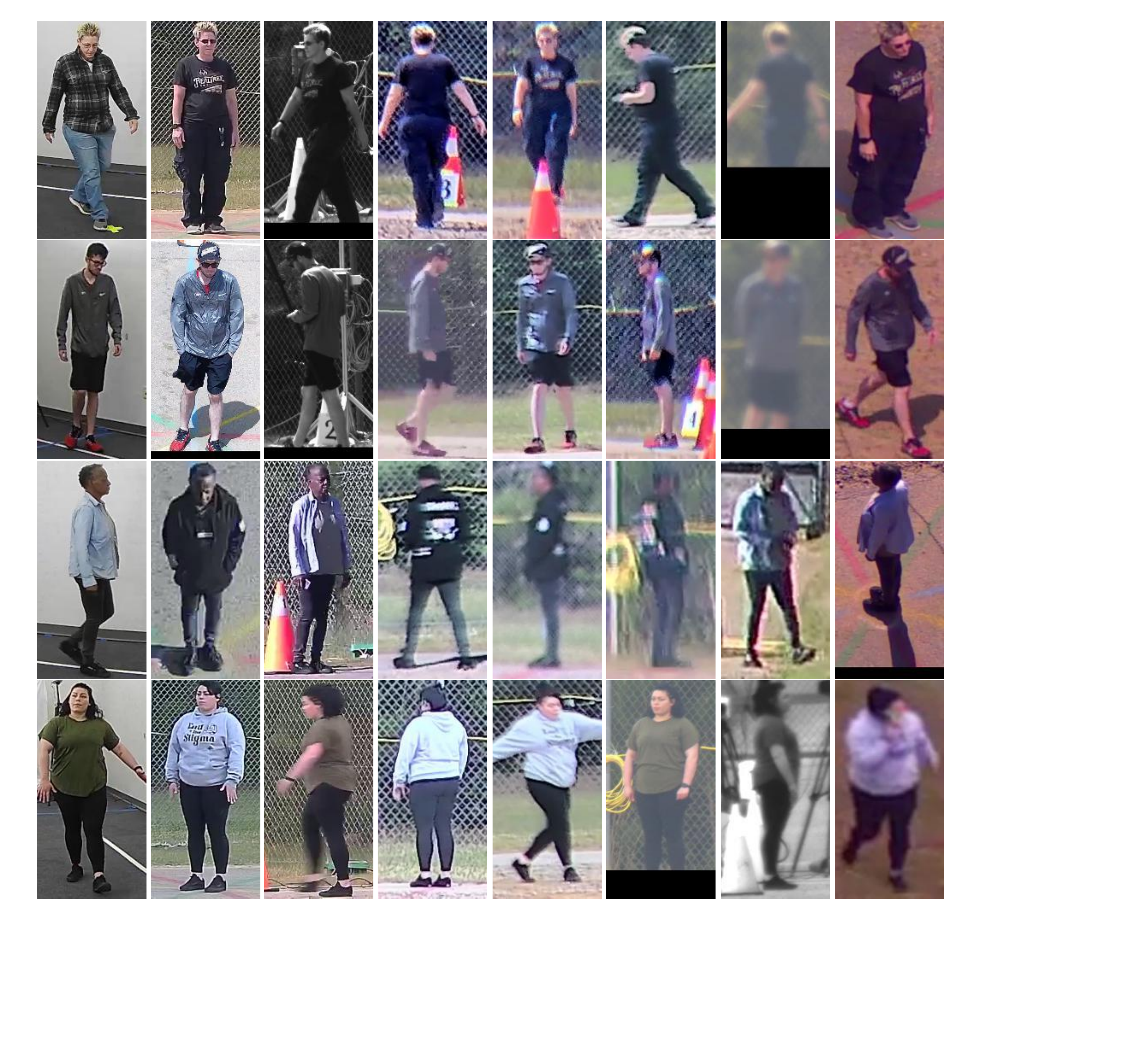}
\end{center}
\vspace{-6mm}
   \caption{\small Example frames in the BRIAR dataset~\cite{cornett2023expanding} showing the same subject (identity) under various conditions, including different standoff distances, clothing, and image quality due to the turbulence effect. The columns represent different scenarios: controlled conditions, close range, $100$m-set1, $100$m-set2, $200$m, $400$m, $500$m, and UAV capture, respectively.}
\label{fig:briar_examples}
\vspace{-5mm}
\end{figure}

The FarSight system faces distinct challenges. 
Captured videos often suffer from poor quality due to long-range capture and atmospheric turbulence.
Elevated platforms introduce large yaw and pitch angles, making data analysis more challenging. 
Extracting identity features is affected by low visual quality, and the training data's limited domain further complicates the learning task. 
Further, the lack of transparency in deep learning models poses a significant issue. Fig.~\ref{fig:briar_examples} illustrates these challenges with examples from close-range, mid-range ($100$-$500$m), and UAV-captured scenarios.

\SubSection{Physics Modeling of Turbulence}

\begin{figure}[t]
\centering
\includegraphics[width=\linewidth]{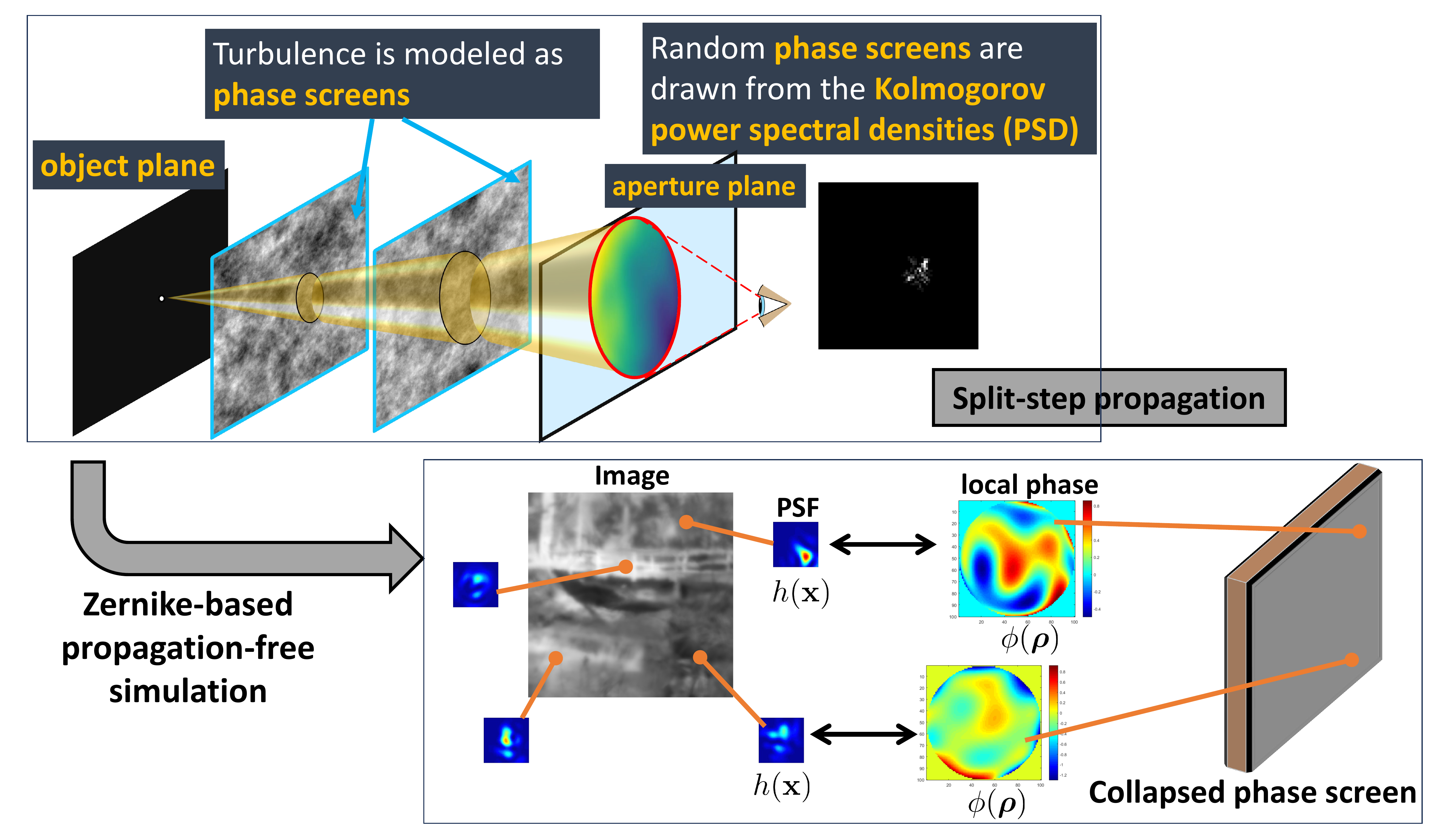}
\vspace{-6mm}
\caption{\small Turbulence modeling. Comparing split-step~\cite{hardie2017simulation,bos2012technique} and Zernike-based simulations \cite{Chimitt_2020_a, Mao_2021_a, chimitt_2022_a}.}
\label{fig: turbulence simulation}
\vspace{-4mm}
\end{figure}

\noindent Atmospheric turbulence is an unavoidable degradation when imaging at range. It is often computationally modeled by splitting the continuous propagation paths into segments via phase screens as illustrated in Fig.~\ref{fig: turbulence simulation}. 
While accurate, the spatially varying nature of the propagation
makes this a computationally demanding process \cite{hardie2017simulation, Hardie_2022_a, Schmidt_2010_a}.

More recent works have explored the possibility of \emph{propagation-free} models where the turbulence effects are implemented as random sampling at the \emph{aperture} \cite{Chimitt_2020_a, Mao_2021_a, chimitt_2022_a}. As shown in Fig.~\ref{fig: turbulence simulation}, every pixel on the aperture is associated with a random phase function which has a linear representation using the Zernike polynomials \cite{Noll_1976_a}. By constructing the covariance matrix of the random process, we can draw samples of the Zernike coefficients to enforce spatial and modal correlations. Propagation-free simulation has enabled $1000 \times $ speed up compared to the split-step propagation methods while maintaining accuracy. Therefore, we adopt this simulation approach in our system.

For the generation of training data, realistic optical and turbulence parameters significantly influence the appearance of the generated defects. Therefore, our datasets are synthesized according to the metadata of various long-range optical systems. Our training dataset also consists of both dynamic and static scenes \cite{Safdarnejad_2015_a, Jin_2021_a, zhou2017places}.

\SubSection{Detection and Tracking}

Our detection module, based on~\cite{wan2021bodyface}, uses a two-stage R-CNN detector~\cite{ren2015faster} with a modified ResNet50 backbone to associate face and body bounding boxes~\cite{wan2021bodyface}. This is done using associative embeddings to match faces and bodies, learned via \textbf{pulling} and \textbf{pushing} loss functions~\cite{duan2019centernet}.
The pulling loss brings embeddings of the same subject closer in the presence of intra-subject variations, calculated as body-to-body, face-to-face, and face-to-body pairs. These are combined using a weighted sum of body-to-face loss, and the sum of face-to-face and body-to-body losses.
Pushing loss, in contrast, pushes away bounding boxes assigned to different subjects to account for inter-subject variations. It is divided into three losses between pairs of body boxes, pairs of face boxes, and body-face pairs. These losses are combined by a weighted sum.
The final associative embedding loss used to optimize these embeddings is a weighted sum of the pulling and pushing losses.

The module also predicts ``head hook" coordinates for every subject to improve body and face association. The head hook loss is a weighted sum of the Smooth L1 loss~\cite{girshick2015fast} and a scale-invariant angular loss.
The final association between body and face bounding boxes is based on similarity metrics, including embedding distance, head hook distance, and confidence scores. The RBF kernel is used for both the embedding distance and head hook distance. The confidence scores factor directly into the association loss to mitigate associating low-confidence bounding boxes with high-confidence ones.
Finally, all these metrics are integrated into a final association metric. If a face prediction's maximum similarity score with any body is below a set threshold, it is concluded that the subject's face is not visible.

\SubSection{Image Restoration} 

Image restoration aims to reverse the image formation process, as described by the equation~\cite{Chan_2022_a}
\begin{equation}
    I(\mathbf{x}) = [\mathcal{B} \circ \mathcal{T}] (J(\mathbf{x})),
    \label{eq: turb_img_form}
\end{equation}
where, $\mathcal{T}$ is the tilt operator and $\mathcal{B}$ represents the blur operation, with $J(\mathbf{x})$ and $I(\mathbf{x})$ as the input and output images, indexed by position $\mathbf{x}$, respectively. In this work, we have considered a single-frame image restoration method as well as a multi-frame method, both aiming to invert $\mathcal{T}$ and $\mathcal{B}$.

Our restoration methods for biometrics focus on preserving identity, using lightweight, real-time techniques. These are divided into single-frame and multi-frame restorations. The former provides lower throughput but relies on strong priors without altering the subject's identity. Multi-frame restoration, on the other hand, utilizes temporal cues, allowing weaker priors but requiring larger throughput.

\begin{figure}[t]
    \centering
    \includegraphics[width = 0.9\linewidth]{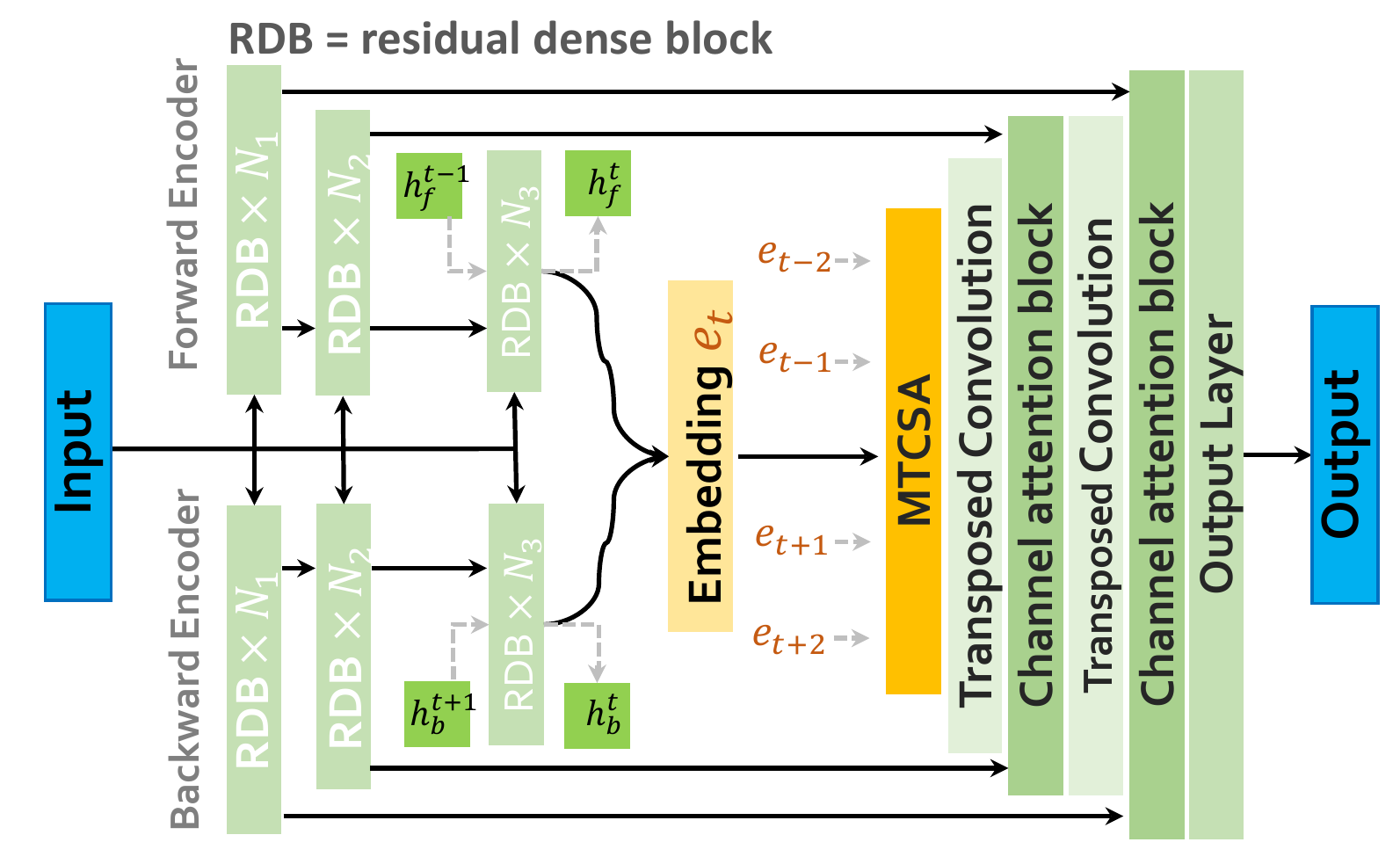}
    \vspace{-3mm}
    \caption{\small Multi-frame image restoration by the recurrent network for turbulence mitigation (RTM).}
    \label{fig: rtm}
\vspace{-4mm}
\end{figure}

Our multi-frame approach uses the Recurrent Turbulence Mitigation network (RTM), a bi-directional, multi-scale convolutional recurrent network with a novel Multi-head Temporal Channel self-attention (MTCSA) layer (Fig.~\ref{fig: rtm}).
%

\SubSection{Multi-Modal Biometric Feature Encoding} 

We describe here our methods for obtaining biometric features from the face, gait and body shape, as well as the multi-modal fusion technique applied to generate fused scores for evaluation on the metrics described in Sec.~\ref{evaluation-metrics}.
\SubSubSection{Face}
\label{face-feature-encoding}
Our face recognition pipeline integrates the techniques of Adaptive Margin Function (AdaFace~\cite{kim2022adaface}), Cluster and Aggregate (CAFace~\cite{caface}), and Controllable Face Synthesis Model (CFSM~\cite{liu2022cfsm}), addressing the challenge of recognizing faces across variable image qualities and media types.

Initially, AdaFace~\cite{kim2022adaface}, an adaptive loss function strategy, helps manage low-quality face datasets. It adjusts the emphasis on misclassified samples based on image quality, effectively dealing with a wide range of image quality levels.
Next, CAFace~\cite{caface}, a two-stage feature fusion technique, is crucial for integrating features from multiple frames. By grouping inputs to a few global cluster centers and subsequently fusing these features, CAFace maintains order invariance while combining multiple frames.
Lastly, CFSM~\cite{liu2022cfsm} helps bridge domain gaps between training and testing scenarios. It replicates the target datasets' distribution in a style latent space, generating synthetic face images similar to the target evaluation datasets, thereby reconciling the disparity between high-quality training data and lower-quality surveillance images.
The combination of AdaFace, CAFace, and CFSM effectively navigates the challenges of face recognition across diverse image qualities, leveraging feature extraction, feature integration, and synthetic image generation to improve face recognition performance.

\SubSubSection{Gait}

We propose an innovative framework, GlobalGait, to address the limitations of existing gait recognition models that mainly focus on local features and often overlook vital global correlations. GlobalGait enriches these local features by factoring in global correlations across a gait sequence, thereby boosting recognition accuracy.

Given an input sequence, GlobalGait uses a CNN backbone to extract local spatiotemporal features, and then divides them into source and target features. These feature maps are projected into tokens for each joint, using sampling around each $2$D joint. We employ a stack of multi-head self-attention layers to model the sequences' spatial and temporal correlations.
Further, GlobalGait attempts to reconstruct target frame pixels based on source sequences and to choose the correct target sequence from a set of candidates. 
This approach harnesses the spatial and temporal correlations in gait recognition, with these supervisory signals guiding the model to learn more distinct gait features.

\begin{figure}[t]
  \centering
  \includegraphics[width=1\linewidth]{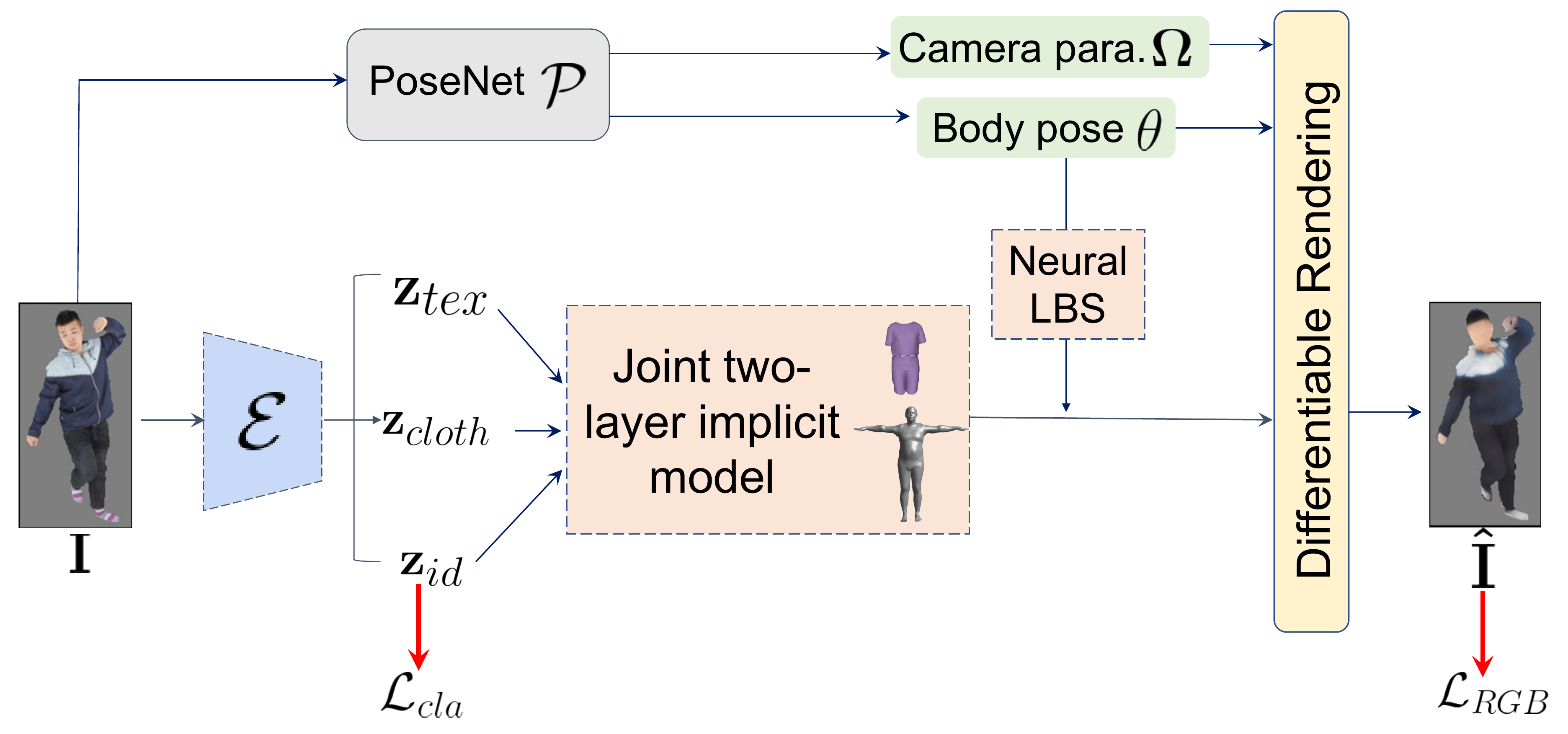}
  \vspace{-7mm}
   \caption{\small Overview of the proposed body shape feature encoding framework (3DInvarReID~\cite{liu2023learning}). In the body matching process, the identity shape features $\mathbf{z}_{id}$ are utilized for matching.}
   \label{fig:bm_flowchart}
   \vspace{-4mm}
\end{figure}

\SubSubSection{Body Shape}
Our method (3DInvarReID~\cite{liu2023learning}) for encoding body features harnesses the power of Person Re-ID~\cite{ye2021deep,ahmed2015improved,zheng2015scalable,li2018harmonious}, with the primary aim to effectively capture static body features. 
We posit that the most reliable cue for body matching is the naked $3$D body shape, despite the considerable challenges in reconstructing it from a $2$D image.
Taking cues from advancements in $3$D feature learning, we introduce a pipeline to disentangle identity (naked body) from non-identity components (pose, clothing shape and texture) of $3$D clothed humans. The core of our approach lies in a novel joint two-layer neural implicit function that disentangles these components in latent representations.

As illustrated in Fig.~\ref{fig:bm_flowchart}, 
%
given a training set of $T$ images $\{\mathbf{I}_i\}_{i=1}^{T}$ and the corresponding identity labels $\{l_i\}_{i=1}^{T}$,
the image encoder $\mathcal{E}(\mathbf{I}):\mathbf{I}\xrightarrow{} (\mathbf{z}_{id}, \mathbf{z}_{cloth}, \mathbf{z}_{tex}$) predicts the identity shape code of naked body $\mathbf{z}_{id}\in\mathbb{R}^{L_{id}}$, clothed shape code $\mathbf{z}_{cloth}\in\mathbb{R}^{L_{cloth}}$ and texture code $\mathbf{z}_{tex}\in\mathbb{R}^{L_{tex}}$. A joint two-layer implicit model decodes the latent codes to identity shape, clothing shape, and texture components, respectively. 
Additionally, PoseNet $\mathcal{P}$ predicts the camera projection $\mathbf{\Omega}$ and SMPL body pose $\mathbf{\theta}$: $(\mathbf{\Omega}, \mathbf{\theta}) = \mathcal{P}(\mathbf{I})$.
Mathematically, the learning objective is defined as:
\begin{equation}
\label{eqn:objective}
\argmin_{\mathcal{E},\mathcal{F},\mathcal{C},\mathcal{T}} \sum_{i=1}^T \left(\left|\hat{\mathbf{I}}_i -  \mathbf{I}_i\right|_1 
+ \mathcal{L}_{cla}(\mathbf{z}_{id},l_i)\right),
\end{equation}
where $\mathcal{L}_{cla}$ is the classification loss. $\hat{\mathbf{I}}$ is the rendered image.
This objective enables us to jointly learn accurate $3$D clothed shape and discriminative shape for the naked body.

We utilize CAPE~\cite{ma2020learning} and THuman2.0~\cite{zheng2019deephuman} datasets to train our model, generating individual identity shape code, clothing shape code, and texture code for each training sample. For inference, the encoder processes body images to extract identity shape features $\mathbf{z}_{id}$. The Cosine similarity of two $\mathbf{z}_{id}$ determines if two images belong to the same person. This method, excluding the explicit $3$D reconstruction during inference, is highly efficient.

\SubSubSection{Multi-Modal Biometric Fusion}
To produce a comprehensive probe-gallery score from multiple biometric modalities, we initially calculate per-modality scores for each probe-gallery pair. For the face, gait, and body, we create a singular subject-level feature using CAFace (Sec.~\ref{face-feature-encoding}), mean fusion on video-only gallery features, and mean fusion on whole-body media, excluding face-only images, respectively. This exclusion is necessary due to the prevalence of face-only gallery images and the unsuitability of gait recognition on single images. 
%
Probe features are then compared to gallery features, and an equal-weighted sum score fusion is employed to generate a single score from the cosine similarity scores of the three modalities. When feature extraction fails for one or more modalities, we impute missing scores to the middle of the score range, which is zero for the cosine similarity metric used in generating probe-gallery scores.
This imputation method was chosen after evaluating alternative techniques, with this approach showing the least bias and greatest stability. 

%% file: sec_4_exp.tex
\begin{table*}[t]
    \centering
\resizebox{0.90\linewidth}{!}{
    \begin{tabular}{|c||p{2.0cm}<{\centering}|p{1.8cm}<{\centering}||p{2.0cm}<{\centering}|p{1.8cm}<{\centering}||p{2.0cm}<{\centering}|p{1.8cm}<{\centering}|}
     \hline
      \multirow{2}{*}{Method} & \multicolumn{2}{c||}{ \makecell{\textbf{Verification (1:1)} \\ TAR@1\% FAR $\uparrow$} } & \multicolumn{2}{c||}{\makecell{\textbf{Rank Retrieval (1:N)} \\ Rank-20, Closed Search  $\uparrow$}}  & \multicolumn{2}{c|}{\makecell{\textbf{Open Search (1:N)} \\ FNIR@1\% FPIR $\downarrow$} } \\
      \cline{2-7}
      & FaceRestricted & FaceIncluded & FaceRestricted & FaceIncluded & FaceRestricted & FaceIncluded \\ 
      
     \hline\hline
 Baseline-AdaFace~\cite{kim2022adaface}     & 
 $9.61$ &  $66.20$ & $14.97$ & $73.85$ & $96.22$ & $70.64$ \\       
 \textbf{FarSight (Face)}    & 
 $25.04$ & $78.01$ & $31.78$ & $84.12$ & $92.11$  & $57.39$\\     \hline  
 Baseline-GaitBase~\cite{gaitbase}    & $44.33$ & $45.55$ & $64.90$ & $68.03$ & $98.53$ & $98.79$ \\       
 \textbf{FarSight (Gait)}    & $56.23$ & $59.20$ & $72.55$ & $74.64$ & $95.24$ & $95.31$  \\   
 \hline  
 Baseline-CAL~\cite{CAL}         & $48.58$ & $51.87$ & $66.27$ & $71.18$ & $96.98$ & $96.17$\\       
 \textbf{FarSight (Body)}    & $51.02$ & $54.00$ & $69.18$ & $72.91$ & $96.95$ & $96.23$ \\     \hline
 \textbf{FarSight (Face+Gait)}     & $57.30$ & $83.98$  & $75.15$ & $91.19$ & $\textbf{87.64}$ & $\textbf{54.55}$ \\
\textbf{FarSight (Face+Body)}     & $54.68$ & $\textbf{85.93}$ & $73.97$ & $\textbf{93.13}$ & $89.57$ & $58.99$ \\ 
 \textbf{FarSight (Gait+Body)}     & $58.91$ & $62.08$ & $73.06$ & $75.57$ & $94.86$ & $94.74$\\ \hline \hline
  AdaFace+GaitBase+CAL      & $51.70$ & $69.15$  & $65.57$ & $80.19$ & $94.92$ & $67.53$\\ 
   \textbf{FarSight}     & $\textbf{63.00}$ & $81.88$ & $\textbf{77.39}$  & $91.74$ & $90.66$ & $67.77$ \\  
 \hline
    \end{tabular}
    }
    \vspace{-2mm}
        \caption{Whole body biometric recognition results on the BRIAR dataset (N=$644$ in retrieval and $544$ in open-set search).}
	\label{tab:bts_results}
 \vspace{-4mm}
\end{table*}

\begin{table}[t]
    \centering
\resizebox{0.98\linewidth}{!}{
    \begin{tabular}{|c|c|c|c|c|c|c|}
     \hline
     Probe & Close range & 100m & 200m & 400m & 500m & UAV \\ 
     \hline\hline
 {FarSight (Face)}    & $68.57$ & $66.07$ & $89.47$ & $90.78$ & $86.32$  & $72.51$ \\       
 {FarSight (Gait)}    & $75.25$ & $73.49$ & $76.53$ & $74.23$ & $71.41$  & $72.89$ \\       
 {FarSight (Body)}    & $72.68$ & $73.25$ & $75.79$ & $77.40$ & $73.91$ & $73.90$ \\    
 {FarSight}         & $88.55$ & $88.01$ & $93.26$ & $93.92$ & $91.81$ & $88.15$ \\
 \hline
    \end{tabular}
    }
    \vspace{-2mm}
        \caption{Rank-20 (\%) on BRIAR at different altitudes and ranges.}
	\label{tab:bts_distance_results}
 \vspace{-3mm}
\end{table}

\section{Experimental Results}\label{sec:exp} 




All modules are run together in a configurable container environment on PyTorch version 1.13.1. We perform experiments on $8$ Nvidia RTX A6000s, with $48$ GiB of VRAM, over the course of $48$ hours on $2$ dual-socket servers with either AMD EPYC $7713$ $64$-Core or Intel Xeon Silver $4314$ $32$-Core processors.

\Paragraph{BRIAR Datasets~\footnote{All human data is collected in accordance with ethical standards and received approval from IRB.} and Protocols.}
%
The IARPA BRIAR dataset~\cite{cornett2023expanding}, comprises two collections—BRIAR Government Collections 1 (BGC1) and 2 (BGC2), is a pioneering initiative to support whole-body biometric research. It addresses the necessity for broader and richer data repositories for training and evaluating biometric systems in challenging scenarios. 
BRIAR consists of over $350,000$ images and $1,300$ hours of videos from $1,055$ subjects in outdoor settings.
The dataset, with its focus on long-range and elevated angle recognition, provides a fertile ground for algorithm development and evaluation in biometrics.

The dataset, in accordance with Protocol V2.0.1, has been partitioned into a training subset (BRS, $411$ subjects) and a testing subset (BTS, $644$ subjects), with non-overlapping subjects.
Regarding the test subjects, we utilize the controlled images and videos as gallery, and the field-collected data as probe. The protocol provides for $644$ subjects for closed-set search and includes two subsets of $544$ subjects each for open-set search, both containing $444$ distractors who lack corresponding probe subjects.
%
%
The probes, totaling $20,432$ templates, are categorized into FaceIncluded and FaceRestricted. FaceIncluded ensures the face is discernible, with at least $20$ pixels in head height. FaceRestricted contains data with challenges like occlusions and low resolution.

\Paragraph{Metrics.}
\label{evaluation-metrics}
We employ BRIAR Program Target Metrics~\cite{bba} to measure  FarSight's performance across multiple modalities and their fusion: 
verification (TAR@1\% FAR), closed-set identification (Rank-20 accuracy), and open-set identification (FNIR@1\% FPIR), allowing for a thorough examination of its performance across various settings. 

\Paragraph{Baselines.}
%
In our study, we utilize established benchmarks for each biometric modality to ensure a comprehensive comparison: For facial recognition, we utilize AdaFace coupled with an average feature aggregation strategy, a popular approach known for its excellent performance~\cite{kim2022adaface}. For gait recognition, we adopt GaitBase~\cite{gaitbase}, a solution known for its efficacy. For body shape modality, we employ CAL~\cite{CAL}, a SoTA cloth-changing person re-identification method. These benchmarks provide an excellent basis to fairly evaluate our proposed method.


\subsection{Evaluation and Analysis}


%

In Tab.~\ref{tab:bts_results}, we provide a thorough comparison of our approaches and the baselines for each modality. The detailed comparison analysis clearly highlights the superior performance of our proposed FarSight system across all performance metrics when compared to the baselines.
For each modality, our module outperforms the baselines by a significant margin. For instance, in the verification metric (TAR@1\% FAR) on FaceIncluded sets, FarSight (Face) sees an increase of $11.81\%$. For gait, there's an improvement of $13.65\%$, and for body shape, we see an improvement of $2.13\%$. Further, upon fusion, we gain an additional improvement of $16.78\%$ ($69.15\%\xrightarrow{}85.93\%$).

The FarSight system's effectiveness across various modalities and distances is evident in Tab.~\ref{tab:bts_distance_results}, displaying each modality's distinct robustness at different ranges. Especially noteworthy is the integrated FarSight model, exhibiting an outstanding accuracy consistently above $88\%$ across all investigated ranges. 
%
The observed increase in face recognition accuracy with distance is tied to the growing similarity between sensors used in training and testing data. As this sensor alignment increases with distance, it reduces the domain gap, leading to enhanced performance. This finding underscores the critical role of sensor type and domain adaptation in optimizing biometric recognition.

%
%

\SubSubSection{Face}
The efficacy of including various modules in the face recognition pipeline is shown in Tab.~\ref{tabl:faceparts}. We initially use the combination of AdaFace IR101 backbone with the average feature aggregation which has shown good performance in low-quality imagery~\cite{kim2022adaface}.  CFSM~\cite{liu2022cfsm} adds performance improvement by adopting training data to a low-quality image dataset WiderFace~\cite{widerface} ($+1.18$ in TAR@1\% FAR). CAFace~\cite{caface} is a feature fusion method that improves upon the basic average pooling (+$4.16$). Lastly, finetuning the model on the BGC1 training dataset further improves the performance (+$6.47$).  
%
%
%
The inclusion of an RTM-based image restoration model, as demonstrated in Table~\ref{tabl:facerestore}, leads to noticeable performance enhancements

\begin{table}[t]
    \centering
    \small
    \resizebox{0.85\linewidth}{!}{
    \begin{tabular}{|c|c|c|c|}
    
    \hline
       FaceIncluded & \makecell{TAR@\\1\% FAR} & \makecell{Rank-\\20} & \makecell{FNIR@\\1\% FPIR} \\ \hline
        AdaFace~\cite{kim2022adaface} & $66.20$ & $73.85$ & $70.64$ \\ \hline
         + CFSM~\cite{liu2022cfsm} & $67.38$ & $77.22$ & $68.51$ \\ \hline
         + CAFace~\cite{caface} & $71.54$ & $78.57$ & $61.77$ \\ \hline
        +BRS1 \textbf{FarSight (Face)}  &  $78.01$	& $84.12$	 & $57.39$   \\ \hline
        
    \end{tabular}}
    \vspace{-2mm}
    \caption{Ablation of different parts in face recognition pipeline. }
    \label{tabl:faceparts}
    \vspace{-2mm}
\end{table}

\begin{table}[t!]
    \centering
    \small
        \resizebox{0.65\linewidth}{!}{
    \begin{tabular}{|r|c|}
    \hline
       TAR@1\% FAR  & \makecell{FaceIncluded} \\ \hline
        Face w/o Restoration  & $72.39$\\ \hline
        Face w/ Restoration   &$\mathbf{72.57}$ \\ \hline
    \end{tabular}}
    \vspace{-2mm}
    \caption{Face recognition with and without image restoration.}
    \label{tabl:facerestore}
    \vspace{-4mm}
\end{table}


\SubSubSection{Gait}
In our gait recognition experiments, we observe consistent improvements compared to GaitBase~\cite{gaitbase}, our baseline, across all four metrics.
Our findings demonstrate significant enhancements in the model's ability to accurately verify individuals, with the TAR@1\% FAR reaching an impressive improvement of \(11.90\%\) in FaceRestricted verification and \(13.65\%\) in FaceIncluded verification. 
Further, the rank-20 metric exhibits notable advancement, showcasing a remarkable increase of \(6.61\%\). 
Lastly, our model showcases improved performance in open-set search, achieving a noteworthy reduction of \(3.29\%\) in FNIR@1\% FPIR.
These promising outcomes reaffirm the efficacy of FarSight (Gait) to extract more discriminative features based on global features and highlight its potential for reliable and robust biometric identification in real-world applications.

\begin{figure}[t]
\begin{center}
\resizebox{1\linewidth}{!}{
\begin{tabular}{r r r r}
{\includegraphics[height=3cm]
{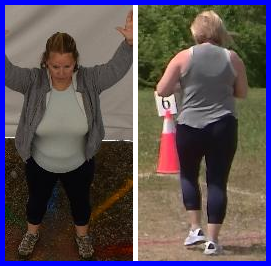}} & 
{\includegraphics[height=3cm]{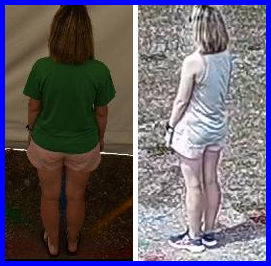}} &
{\includegraphics[height=3cm]{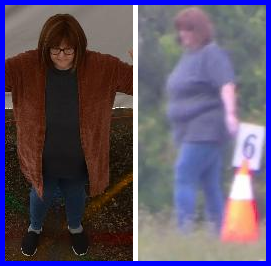}} & \vspace{-1.5mm} 
{\includegraphics[height=3cm]{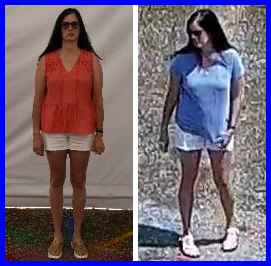}}  \\
\textcolor{red}{close range}& \textcolor{red}{close range} & \textcolor{red}{$400m$} & \textcolor{red}{close range}\\
\multicolumn{4}{c}{\Large (a) Successful recognition of genuine pairs}\vspace{1mm}\\ 
{\includegraphics[height=3cm]{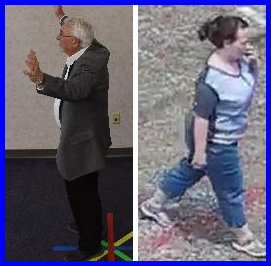}} & 
{\includegraphics[height=3cm]{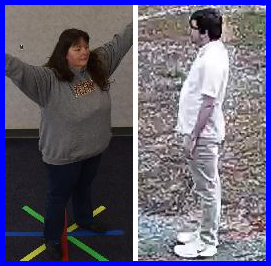}} &
{\includegraphics[height=3cm]{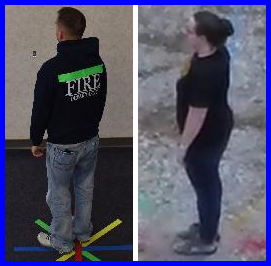}} & \vspace{-1.5mm} 
{\includegraphics[height=3cm]{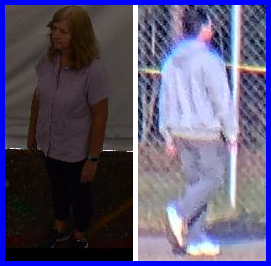}}  \\
\textcolor{red}{close range}& \textcolor{red}{close range} & \textcolor{red}{UAV} & \textcolor{red}{$400m$}\\
\multicolumn{4}{c}{\Large (b) Successful recognition of impostor pairs}\vspace{1mm}\\
{\includegraphics[height=3cm]{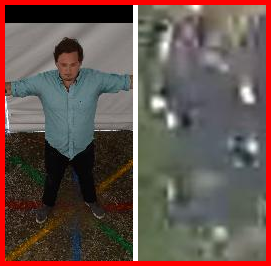}} & 
{\includegraphics[height=3cm]{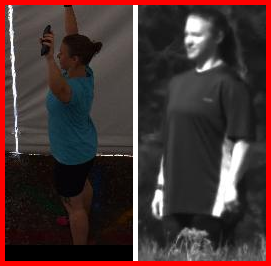}} &
{\includegraphics[height=3cm]{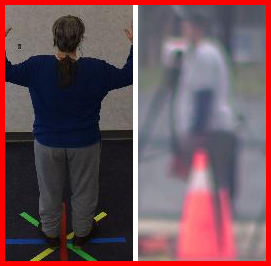}} & \vspace{-1.5mm} 
{\includegraphics[height=3cm]{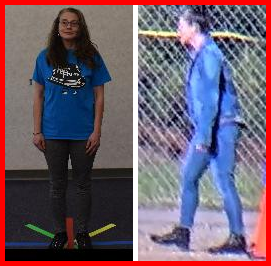}}  \\
\textcolor{red}{$500m$}& \textcolor{red}{$200m$} & \textcolor{red}{$500m$} & \textcolor{red}{$300m$}\\
\multicolumn{4}{c}{\Large (c) Failure cases for genuine pairs}\vspace{1mm}\\
{\includegraphics[height=3cm]{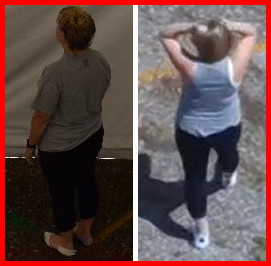}} & 
{\includegraphics[height=3cm]{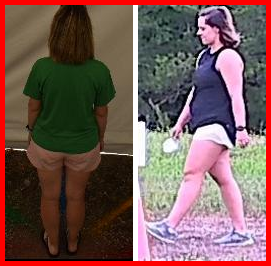}} &
{\includegraphics[height=3cm]{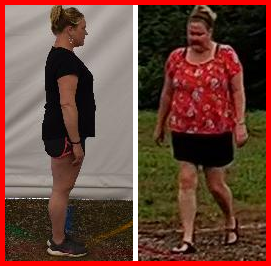}} & \vspace{-1.5mm} 
{\includegraphics[height=3cm]{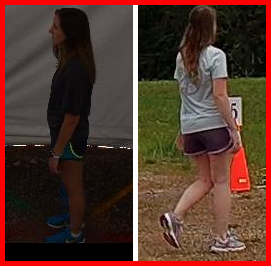}} \\
\textcolor{red}{UAV}& \textcolor{red}{$100m$} & \textcolor{red}{close range} & \textcolor{red}{close range}\\
\multicolumn{4}{c}{\Large (d) Failure cases for impostor pairs}\\
\end{tabular}
}
\end{center}
\vspace{-5mm}
\caption{\small Successful and failure examples of body matching.
} 
\label{fig:body_matching_paired_examples}
\vspace{-2mm}
\end{figure}

\SubSubSection{Body}
Tab.~\ref{tab:bts_results} clearly demonstrates that our FarSight (body) consistently outperforms the CAL baseline on both FaceRestricted and FaceIncluded sets, as evidenced in both verification and Rank retrieval metrics.
In Fig.~\ref{fig:body_matching_paired_examples}, we show successful and failed matches in body matching. Our method copes well with clothing differences, but struggles with motion blur, turbulence, or hairstyle changes. Misidentifications in impostor pairs often happen due to similar body shapes.


\SubSubSection{Multi-Modal Fusion}


As seen in Tab.~\ref{tab:bts_results}, the fusion of three modalities improves over the next best-performing algorithm in the FaceRestricted condition ($+11.30$ in TAR@1\% FAR and $+11.82$ in Rank-20). We also see the strength of combining the face and body modalities in the FaceIncluded condition, where face and body fusion excels in both verification and rank retrieval ($+1.95$ TAR@1\% FAR and $+1.94$ Rank-20) over the next best algorithm. The open search metric performs best when fusing face and gait, scoring $87.64$\% and $54.55$\% in FNIR@1\% FPIR for both the FaceRestricted and FaceIncluded conditions, which is in part due to the challenge that single body and gait modalities on open-set search.


\SubSection{System Efficiency}
 
%
\Paragraph{Template Size.} Feature vectors for face, gait and body are of sizes $512$, $8704$ and $6144$. Multiplying these values by $8$ and dividing by $1024$ provides the template size: $4$KB, $68$KB and $48$KB, respectively, and $120$KB in total.

\Paragraph{Processing Speed.} The speed of our FarSight system, as outlined in Tab.~\ref{tab:processing_time}, is examined under stringent conditions to gauge both the efficiency of individual components and the overall pipeline. 
This system operates asynchronously and concurrently, similar to the actual deployment conditions. 
To precisely measure efficiency, the components are assessed in a serialized manner, even though they typically run in parallel.
We conduct this assessment using representative sample videos, encompassing $2400$ frames of $1080$p and $1200$ frames of $4$K video, each set originating from four distinct subjects. 
The restoration process is primarily directed towards detected faces, which implies that any instances of undetected faces would naturally lead to reduced restoration and face module processing times. A notable observation is that our system can successfully detect bodies in $95$\% of all frames and faces in $26$\% of frames.


\begin{table}
    \centering
    \resizebox{1\linewidth}{!}{
    \begin{tabular}{|c|c|c|c|}\hline
        Module & 1080p & 4K & Average Combined\\ \hline \hline
        Detection \& Tracking & $20.0$ & $34.7$ & $24.9$ \\ \hline
        Restoration           & $6.1$  & $5.3$  & $5.9$\\ \hline
        Face                  & $2.6$  & $2.2$  & $2.5$\\ \hline
        Gait                  & $3.3$  & $2.5$  & $3.0$\\ \hline
        Body                  & $3.7$  & $3.1$  & $3.5$\\ \hline \hline
        FarSight System (fps)& $8.4$ & $6.3$ &  $7.8$\\ \hline
    \end{tabular}
    }
    \vspace{-2mm}
    \caption{\small FarSight module processing times (sec.) and system efficiency (fps) for 1080p (1920x1080) and 4k (3840x2160) probes.}
    \label{tab:processing_time}
    \vspace{-4mm}
\end{table}





%% file: sec_5_conclusion.tex
\section{Future Research}

\Paragraph{Image restoration.} We plan to expand our optical simulation tool to handle higher levels of distortion and explore ``simulation-in-the-loop" techniques. Our goal is also to balance fidelity and perceptual quality by integrating generative and discriminative restoration methods.

\Paragraph{Detection and tracking.} We plan to refine our current detector or shift to YOLO-based detectors. We are also considering using separate face detectors on subject bounding boxes to reduce latency.

\Paragraph{Biometric feature encoding.} 
In our face module, we are exploring the potential of adaptive restoration based on the available information from given frames, to avoid any negative impact on performance. For our gait module, our goal is to delve further into the usage of $3$D body shape and pose information, which is currently under-explored in gait recognition. This involves combining shape parameters with global features to generate $3$D-aware shape features and enriching local features with $3$D pose information. For body analysis, we aim to refine $3$D body reconstructions using multiple frames and assess the value of $3$D poses compared to $2$D imagery. Future research will encompass additional baselines, including face, gait, and body shape.

%
\Paragraph{Multi-modal fusion.} We plan to further enhance our technique for fusing face, gait, and body features, to better exploit the strengths of each modality and alleviate challenges from the long tail of body and gait scores in the non-match open search distributions.

\section{Conclusion}\label{sec:conclusion}

We develop and prototype an end-to-end whole-body person recognition system, \textbf{FarSight}. Our solution attempts to overcome hurdles such as low-quality video frames, large yaw and pitch angles, and the domain gap between training and test sets by utilizing the physics of imaging in harmony with deep learning models. This innovative approach has led to superior recognition performance, as demonstrated in tests using the BRIAR dataset. With the far-reaching potential to enhance homeland security and forensic identification, the FarSight system paves the way for the next generation of biometric recognition in challenging scenarios.

\Paragraph{Acknowledgments.} This research is based upon work supported in part by the Office of the Director of National Intelligence (ODNI), Intelligence Advanced Research Projects Activity (IARPA), via 2022-21102100004. The views and conclusions contained herein are those of
the authors and should not be interpreted as necessarily representing the official policies, either expressed or implied, of ODNI, IARPA, or the U.S. Government. The U.S. Government is authorized to reproduce and distribute reprints for governmental purposes notwithstanding any copyright annotation therein.